%% file: acl_latex.tex
\pdfoutput=1
\documentclass[11pt]{article}

\usepackage[final]{acl}

\usepackage[colorinlistoftodos]{todonotes}

\usepackage{times}
\usepackage{latexsym}
\usepackage{booktabs}
\usepackage[T1]{fontenc}
\usepackage[utf8]{inputenc}

\usepackage{microtype}

\usepackage{inconsolata}
\usepackage{subcaption}

\usepackage{graphicx}
\usepackage{amsmath} % For additional math symbols, if needed
\usepackage{tabularx}
\usepackage{amsmath,amssymb}
\usepackage{algorithm,algpseudocode}
\usepackage{graphicx}
\usepackage{multirow}

\title{Hierarchical Level-Wise News Article Clustering via Multilingual Matryoshka Embeddings}

\author{Hans W.  A. Hanley \\
  Stanford University \\
  \texttt{hhanley@cs.stanford.edu} \\\And
  Zakir Durumeric \\
  Stanford University\\
  \texttt{zakir@cs.stanford.edu} \\}

\begin{document}
\maketitle
\begin{abstract}
Contextual large language model embeddings are increasingly utilized for topic modeling and clustering. However, current methods often scale poorly, rely on opaque similarity metrics, and struggle in multilingual settings. In this work, we present a novel, scalable, interpretable, hierarchical, and multilingual approach to clustering news articles and social media data. To do this, we first train multilingual Matryoshka embeddings that can determine story similarity at varying levels of granularity based on which subset of the dimensions of the embeddings is examined. This embedding model achieves state-of-the-art performance on the SemEval 2022 Task 8 test dataset (Pearson $\rho$ = 0.816). Once trained, we develop an efficient hierarchical clustering algorithm that leverages the hierarchical nature of Matryoshka embeddings to identify unique news stories, narratives, and themes. We conclude by illustrating how our approach can identify and cluster stories, narratives, and overarching themes within real-world news datasets.
\end{abstract}

\input{intro}
\input{related}
\input{datasets}

\input{methodology}

\input{results}
\bibliography{custom}

\appendix

\input{appendix}

%\section{Example Appendix}
%\label{sec:appendix}

%This is an appendix.

\end{document}

%% file: intro.tex
\section{Introduction}

The news ecosystem is increasingly globalized and fractured, with news stories and misinformation spreading across thousands of news websites and between countries' news ecosystems~\cite{rupnik2016news}. For example, during the Russian invasion of Ukraine, a propaganda story about the Nazi composition of the Ukrainian government published in Russia was covered extensively elsewhere --- not just in Russia and Ukraine, but also in the US, China, and throughout the Global South~\cite{blank2022russia,hanley2023happenstance}. Understanding how media outlets cover events is crucial for identifying biases in news platforms, gaps in coverage, and how news ecosystems interact and influence each other~\cite{hanley2025tracking,bisandu2018clustering}. 

Despite the importance of understanding news coverage in a multilingual and global setting~\cite{chen2022semeval,bisandu2018clustering}, most current approaches that use LLM-based text embeddings are monolingual, do no not scale, or are unable to differentiate between news stories and narratives at varying levels of granularity (\textit{i.e.}, identify different overarching themes, topics, and individual narratives)~\cite{hanley2024specious,chen2022semeval,{grootendorst2022bertopic,nielsen2022mumin,xu2022hfl,abdelrazek2023topic}}. Beyond decoder-based large language models like OpenAI's GPT-4 and Anthropic's Claude, which remain prohibitively expensive for processing documents at scale, encoder-based models often simply identify the ``similarity'' overlap of two news articles via the cosine-similarity of their embeddings. Given that this similarity is usually weakly defined, identifying larger themes among documents with encoder-based models remains challenging~\cite{chen2022semeval,chambers2009unsupervised}. Furthermore, even with highly interpretable and robust embeddings, clustering-based approaches to extract out different individual stories are often difficult to utilize because the number of news stories, topics, and themes reported is not known \textit{a priori}~\cite{monath2021scalable,hanley2024specious,jiang2012small,dinari2022revisiting}. As a result, developing algorithms that can automatically determine the number of news stories, topics, and themes within a dataset is essential, unlike traditional approaches such as LDA and K-means~\cite{jelodar2019latent,ahmed2020k}.

To address these challenges, in this work, we introduce a novel adaptation of multilingual hierarchical Matryoshka embeddings and a hierarchical clustering algorithm adapted for clustering semantically similar news articles. Our multilingual Matryoshka embeddings progressively learn additional detail in their upper dimensions, allowing for the identification and differentiation of multilingual news articles at varying granularities of similarity. In our approach, the upper dimensions of embeddings are used to determine if two news articles are about the same event, the middle dimensions to determine if they address the same topic, and the lower dimensions to determine if they address the same theme. Compared to traditional embeddings, our approach makes similarity calculations at varying levels more interpretable while also reducing the cost of similarity calculations. To then identify distinct news stories, topics, and themes, we develop a novel agglomerative clustering algorithm that leverages the naturally hierarchical structure of our Matryoshka 
 embeddings based on the Reciprocal Nearest Neighbor algorithm~\cite{monath2023online}. Our contributions are thus:

\begin{itemize}
\item The design of a multilingual Matryoshka embedding model capable of differentiating between news articles at varying levels of semantic similarity that achieves state-of-the-art performance on the SemEval 2022 Task 8 test dataset (Pearson $\rho=0.816$). 

\item  The design of a hierarchical agglomerative clustering algorithm that exploits the hierarchical nature of our Matryoshka embeddings and a model capable of providing human-interpretable summaries of clusters. 

\end{itemize}

\noindent We release the weights and synthetic portions of our training datasets at \url{https://github.com/hanshanley/multilingual-matryoshka-news}.

\noindent
%Oure models, embeddings, and algorithms will enable the identification of news stories, topics, and themes across different languages, contributing to a deeper understanding of the globalized news ecosystem.

%% file: related.tex
\section{Background and Related Work\label{sec:related}}

\textbf{Semantic Embeddings.} Recent work has investigated models that encode texts into fixed-length embeddings that capture their semantic and syntactic meaning~\cite{reimers2019sentence,li20242d,cer2018universal}. These embeddings are often used for classification, clustering~\cite{hanley2024specious}, semantic textual similarity~\cite{gao2021simcse}, and retrieval-augmented generation~\cite{gao2023retrieval}. While initial work focused on word embeddings~\cite{mikolov2013distributed, pennington2014glove}, recent investigations have concentrated on sentence, paragraph, and document-level representations~\cite{gao2021simcse,li20242d}.

Several recent studies have utilized contrastive learning techniques to achieve state-of-the-art results in designing both monolingual and multilingual semantic embeddings~\cite{gao2021simcse}. Contrastive learning objectives generally seek to bring items that have the same class or label close to each other in the embedding space while simultaneously distancing items that have different classes or labels. For example, \citet{gao2021simcse} use a contrastive loss objective, dropout, and natural language inference to learn monolingual embeddings on top of BERT and RoBERTa-based models. In a similar vein, \citet{wang2022text} utilize a dataset of text pairs and a student-teacher model to train high-quality embeddings.

%Beyond monolingual embeddings, \citet{wang2022text} adopt synthetic data created by GPT3.5/4~\cite{achiam2023gpt} as weakly supervised contrastive learning utilizing multilingual text pairs to extend the E5 model to a multilingual setting of 93~languages. Finally, \citet{park2024improving} and~\citet{zeng2023soft} use contrastive learning and language-ware training methods to extend models to a multilingual setting. 

\noindent
\textbf{Matryoshka Embeddings.} Matryoshka representation learning (MRL)~\cite{kusupati2022matryoshka} is a recent embedding method that seeks to learn flexible nested representations. Namely, MRL optimizes the original loss function (\textit{i.e.}, contrastive loss) at $O(log(d))$ different representation sizes of the full embedding size $d$. Given a labeled dataset \( D = \{(x_1, y_1), \dots, (x_N, y_N)\} \),  where $\mathbf{H}$ are the corresponding embeddings/d-dimensional representations of the datapoints $\mathbf{X}$, and $\mathcal{L}(\cdot)$ is the original function loss, MRL optimizes the loss across several nested truncated versions of the embedding. As shown by~\citet{kusupati2022matryoshka}, this advancement enables more compact representations but comparably accurate representations of semantic information for use cases when the storage size or the flexibility of embeddings is important. 

%\noindent
%For example, when training an embedding model of dimensionality 768, which often uses self-supervised contrastive loss between different texts or items, the contrastive loss would be applied at the dimensions 768, 384, 192, etc. 
\noindent
\textbf{Clustering.} Clustering is an unsupervised algorithm that seeks to identify groups of similar items in a dataset~\cite{bateni2024s}.  Superficially, clustering consists of creating  partition $\mathcal{S} = \{C_1, \cdots, C_k\}$ of a dataset $\mathcal{X} = \{x_i\}_{i=1}^N$, such that $\bigcup_{C \in \mathcal{S}} C = \mathcal{X}$ and $\forall C, C' \in \mathcal{C}, C \cap C' = \emptyset$. Clustering has been used for feature extraction~\cite{zhang2006unsupervised}, visualization~\cite{schwartz2020toomanycells}, and topic analysis~\cite{grootendorst2022bertopic}. 
In this work, we modify a hierarchical clustering algorithm called Reciprocal Agglomerative Clustering (RAC) to make use of Matryoshka embeddings. We give a brief overview here: 

\noindent
\textit{Reciprocal Agglomerative Clustering (RAC).} Starting from all data points, RAC progressively merges clusters in distinct rounds if, among the current set, clusters are each other's reciprocal nearest neighbor (RNN)~\cite{sumengen2021scaling}. While requiring the computation of the distance matrix between clusters, which can be prohibitive, as shown by~\citet{sumengen2021scaling}, since clusters can be joined together in any particular order without affecting cluster quality, this algorithm is easily parallelizable. See ~\citet{sumengen2021scaling} for details.

%% file: datasets.tex
\section{Dataset}

We primarily utilize an expanded version of the SemEval-2022 Task 8: Multilingual news article similarity dataset (SE-22-t8)~\cite{chen2022semeval,chen2024multilingual}. This dataset consists of pairs of news articles from early 2020 that are graded across varying aspects of similarity, including their geographical focus, their narrative schemes, the time period in which they were written, and their tone. We specifically use \textsc{overall} similarity specified in this dataset, which grades news article pairs on whether ``the two articles cover the same substantive news story (excluding style, framing, and tone).'' The \textsc{overall} metric is based on human participants grading article pairs as being ``Very Similar'', ``Somewhat Similar'', ``Somewhat Dissimilar'', and ``Very Dissimilar.'' For example, ``Very Similar'' news articles are about the same news event. We supplement this dataset with \citet{miranda2018multilingual}'s news clustering dataset of multilingual articles. For this work, we treat articles about the same event as defined by~\citet{miranda2018multilingual} as ``Very Similar.'' We provide examples of article pairs of varying similarity in Appendix~\ref{sec:pair-examples}; additional examples and the code book for grading article similarity can be found at \url{https://zenodo.org/records/6507872}~\cite{chen2022semeval,chen2024multilingual}.

%\footnote{\url{https://zenodo.org/records/6507872}}
Given that the published dataset does not include article text, we scraped each of the 37,394~URLs provided. After removing dead links, unavailable articles, and articles shorter than 500~characters, we were left with 24,871~articles in English, Chinese, Spanish, German, Arabic, Italian, Turkish, Polish, French, and Russian. As recommended by~\citet{chen2022semeval}, we utilize the \texttt{newspaper3k} Python library to extract article contents and use the \texttt{BeautifulSoup} library to remove any HTML tags, JavaScript, URL hyperlinks, as well as newline (\verb|\n|) and tab characters (\verb|\t|).

\noindent
\textbf{Dataset Augmentations.}
After scraping the set of news articles from the expanded SE-22-t8 dataset, we augment the dataset with synthetic writing style alterations, entity replacements, and extend the dataset beyond the original 10~languages. 

\noindent
\textit{Stylistic Augmentations.} To ensure that the dataset contains multiple stylistic variations of individual articles to enable our final trained embeddings to recognize that different writing styles about the same event, we use the Open AI model GPT-4o\footnote{We query GPT-4o (\url{https://openai.com/index/hello-gpt-4o/}) given its multilingual capabilities and the reduced cost per token.} to rewrite each article in the original dataset three times with a different stylistic variation using the prompt: ``\textit{Rewrite, in the article's own language (e.g., Spanish, English, Finnish, etc...), the following news article with a different style as if it came from a different website while maintaining the original meaning of the article}.'' See Appendix~\ref{sec:appendix-rewritten} for examples.

\noindent
\textit{Entity Sensitivity.} We ensure that our final models are sensitive to the named entities within different texts, regardless of stylistic similarities between articles. To do this, as in~\citet{mitchell2023detectgpt} and~\citet{hanley2023machine}, we utilize Spacy\footnote{
\url{https://spacy.io/}} to identify named entities in 10,000~articles in the English subsection of the SemEval dataset and subsequently utilize a T5-based model \texttt{T5-1.1-XL} to replace these entities with generic replacements based on the surrounding text. As argued by~\citet{mitchell2023detectgpt}, genetic language models like T5 can create ``meaningful variations'' upon replacing parts of the text. In line with the definitions within the SemEval 22 dataset, we treat these variations as ``Somewhat Similar'' to their original versions. Upon replacing these entities, we additionally translate 500 of these texts to our 53 other leagues using GPT-4o. Altogether, we add an additional 27,000 examples to our dataset.

\noindent
\textit{Additional Languages.} To augment the SE-22-t8 dataset beyond the original 10~languages, for a random selection of article pairs in the original dataset, we query GPT-4o to rewrite the articles to a random subset of 28~additional languages from a set of 54~total languages including Latvian, Chinese, Japanese, and Albanian (full list in  Appendix~\ref{sec:languages}).\footnote{\url{https://help.openai.com/en/articles/8357869-how-to-change-your-language-setting-in-chatgpt}} We utilize the following prompt: ``\textit{Translate the following article into  \{language\}}.''

After querying GPT-4o to rewrite and translate individual article pairs, we take different combinations of translations of the original dataset to increase the size of our dataset. This is such that, for example, after gathering the 29~translations (28~queried + original article) of each article, we extend the original article pair from one data point to 841~data points (\textit{i.e.}, each data point can be paired with each of the different translations of the opposing data point). Similarly, after translation, where each article has 28~translations, we further augment our dataset by gathering each translation pair and treating them as a new pair of ``Very Similar'' articles. Altogether, after rewriting and translating the original set of articles, removing any errors returned by GPT-4o, we extend the original dataset to 4.10M~article pairs. We utilize 10\% (410K article pairs) of our SE-22-t8 dataset as validation.

\vspace{3pt}
\noindent
\textbf{Test Dataset.}
For our evaluation, in addition to the original 10-language SE-22-t8 ~\cite{chen2022semeval} {test} split of 3,958 article pairs and 7,842 unique articles, for each of the 54~languages in our dataset, we translate 1,300~random test instances.  Altogether, these translations enable us to extend the original {test} split to 2.07M~examples. In addition to extending our test set to include languages previously not within the original SE-22-t8 test dataset, this extension enables us to further test and ensure the isomorphism~\cite{marchisio2022isovec} (\textit{i.e.}, that the resulting embeddings are language-agnostic) of our embeddings. Namely, during the evaluation phase of our work, we estimate the relational similarity between each non-English language and English.  See the Appendix~\ref{sec:isomorphism} for details about the calculation of relational similarity.

%we translate all test instances into English and calculate the relational similarity that each language with English.

%% file: methodology.tex
\section{Methodology\label{sec:methodology}}
Our approach operates in two steps. First, we utilize the naturally hierarchical nature of Matryoshka representation learning (MRL) to train contextual embeddings that model the relationship from general textual concepts down to specific descriptions of particular events. Second, we use these embeddings in a level-wise hierarchical clustering algorithm~\cite{monath2023online} to identify specific news stories, topics, and themes. While several previous approaches have sought to build and label concept hierarchies utilizing embeddings~\cite{le2019inferring, stein2019analysis, meng2020hierarchical}, we propose using MRL learning and a supervised approach to first encode generalized hierarchical information. This allows our clustering algorithm to directly use the embeddings to create a hierarchical representation of a text dataset.

\begin{figure}
  \centering
  \includegraphics[width=0.8\columnwidth]{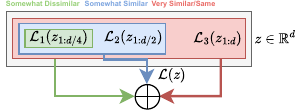}
  \caption{Matryoshka Representation Learning for embedding hierarchical structure into the contextual embeddings of outputted by a semantic encoder. \label{figure:matryoshka}}
  
%This is such that the first $z_{1:d/4}$ embeddings can distinguish between texts that have some similarity (somewhat dissimilar or greater similarity) and texts that are unrelated. Similarly, the first  $z_{1:d/2}$ embeddings can distinguish between texts that are somewhat similar (somewhat similar or greater similarity)and texts that are unrelated. Finally, the full embedding structure can distinguish between embeddings that are about the same or different news stories or events. 
\label{fig:mainstream-over_time}
\vspace{-15pt}

\end{figure}

\noindent
\textbf{Matryoshka Embeddings for Fine-Grained Similarity Calculations.} As previously noted, Matryoshka representation learning (MRL) is a flexible alternative to traditional embedding approaches, which learns embeddings of varying capacities through the explicit optimization of lower-dimensional vectors in a nested fashion~\cite{kusupati2022matryoshka}. However, within this work, rather than learning the same representation at multiple dimensions, we use our Matryoshka embeddings to learn increasingly specific information about news articles by taking advantage of the similarity labels from the SE-22-t8 dataset. %Using the extended SemEval 2022-t8 dataset~\cite{chen2022semeval}, we use MRL to differentiate between articles of increasing levels of similarity at higher embedding dimensionalities, encoding baseline and thematic elements in lower embeddings and higher-level details in the upper embeddings. 

\noindent
\textit{Modified AngIE Loss for MRL.} To train our Matryoshka embeddings, we utilize a modified version of contrastive AngIE loss~\cite{li2023angle}, applying this loss at $O(log(d))$ different representation sizes of the full embedding size $d$ (see Figure~\ref{figure:matryoshka}). We adapt AngIE loss, as it largely depends on the relative ranking between different datapoint pairs for calculating loss, allowing us to directly use the rankings of the similarities (\textit{i.e.}, ``Very Similar'' $>$ ``Very Dissimilar'') in the SE-22-t8 dataset. Specifically,  given a dataset with set of embedding pairs $x_i =(\mathbf{h_m}, \mathbf{h}_n )$ that have labeled similarities $ y_i = s(\mathbf{h}_m, \mathbf{h}_n)$, AngIE loss consists the sum of a cosine objective function, (2) a contrastive in-batch negative objective (2), and finally (3) an angle objective:

{\small
\begin{equation*}
 \mathcal{L}_{\text{AngIe}} = \mathcal{L}_{\text{cos}}+\mathcal{L}_{\text{contrast}}+\mathcal{L}_{\text{angle}}
\end{equation*}
\begin{equation*}
\mathcal{L}_{\text{cos}} = \log \left[ 1 + 
\sum_{\substack{s(\mathbf{h}_i, \mathbf{h}_j) > \\ s(\mathbf{h}_m, \mathbf{h}_n)}} 
\exp \left( 
\frac{\cos(\mathbf{h}_m, \mathbf{h}_n) - \cos(\mathbf{h}_i, \mathbf{h}_j)}{\tau} 
\right) 
\right]
\end{equation*}
\begin{equation*}
  \mathcal{L}_{\text{contrast}} = \sum_i^{N}  -\log \frac{\sum_{k+}^{} e^{\cos(h_i, h_k^+)/\tau}}{\sum_{j=1}^N e^{\cos (h_i, h_j)/\tau}}  
\end{equation*}
\begin{equation*}
\mathcal{L}_{\text{angle}} = \log \left[ 1 + 
\sum_{\substack{s(\mathbf{h}_i, \mathbf{h}_j) > \\ s(\mathbf{h}_m, \mathbf{h}_n)}} 
\exp \left( \frac{\Delta \theta_{ij} - \Delta \theta_{mn}}{\tau} \right) 
\right]
\end{equation*}

}
\noindent
where $\tau$ is the temperature hyperparameter, $cos(\cdot,\cdot)$ is the cosine similarity function, and $(\mathbf{h_i}, \mathbf{h}_k^+ )$ within a batch are pairs that have the same class, label, or are considered to be positives of high similarity. For details on how the complex representations of the contextual text representations are computed and how the normalized angle differences are determined, see Appendix~\ref{sec:app-angle-obj} or~\citet{li2023angle}.

To adapt the AngIE loss to the MRL setting and to identify multilingual articles of increasing similarity, we make two key changes to the AngIE objective: (1) at progressively higher dimensions, we increase the threshold for considering a news article pair as similar or as positives and (2) we incorporate SimCSE~\cite{gao2021simcse} for improved monolingual embedding spaces. This is such that for the loss applied at $d/4$-dimensions we treat ``Very Dissimilar'' pairs as having a labeled cosine similarity of 0, and all other pairs as having a labeled cosine similarity of 1; then for the loss applied at $d/2$ dimensions,  we treat ``Very Dissimilar'' and ``Somewhat Dissimilar'' pairs as having a labeled cosine similarity of 0, and all other pairs as having a labeled cosine similarity of 1, \textit{etc...} This forces the embedding to progressively learn to differentiate different levels of similarity, utilizing the lower dimensions to learn higher concepts and later dimensions to learn finer-grained details.  

To incorporate  SimCSE~\cite{gao2021simcse}, during training, we encode each batch twice through our encoder with different dropout masks. As found by~\citet{gao2021simcse}, utilizing identical positive pairs that have independently sampled dropout masks by feeding the same input to the encoder twice can lead to higher quality embeddings. Given that we are training multilingual embeddings, this additional step to explicitly train each monolingual training instance with itself further leads to better individual language embedding spaces~\cite{park2024improving}. Our final MRL loss is thus:
\begin{align*}
  \mathcal{L}_{mat} &=  \mathcal{L}_{\text{AngIE}_{diss}}(\mathbf{H}_{d/4}) \nonumber \\
  &\quad + \mathcal{L}_{\text{AngIE}_{somewhat}}(\mathbf{H}_{d/2}) + \mathcal{L}_{\text{AngIE}_{same}}(\mathbf{H}_{d})
\end{align*}

\noindent
%\textit{Further Multilingual Alignment.}
%While when initially training our Matryoshka Embeddings, our objective \emph{does} pull translations of individual articles closer to each other in the embedding space, to further ensure that our model is language-agnostic and isomorphic between languages, we employ an additional post-training student-teacher approach to ensure that our model is properly aligned between different languages~\cite{reimers2020making}. This ensures that when clustered, our model will not lead to separate clusters for articles that convey the same story for different languages. To do this, we take our original Matryoshka model $M$ as well as a multilingual text pair $(s_i, t_i)$ where $s_i$ is in English and  $t_i$ could be in any language. We then as in ~\citet{reimers2020making}, subsequently train a student contextual encoder $\hat{M}$ (initialized with weights of $M$), such that $\hat{M}$ such that cos($\mathbf{h}_{\hat{M}(t_i)}, \mathbf{h}_{M({s_i})}) \approx 1$. For a given batch of size N, we then minimize the mean-squared loss:

%{\small
%\begin{align*}
%\frac{1}{N} \sum_{i=1}^N   \left(1- \text{cos}%(\mathbf{h}_{M(s_i)} - \mathbf{h}_{\hat{M}(t_i)}) \right)^2 
%\end{align*}
%}

%\noindent
%We apply this loss function at each of the different dimensions required by our Matryoshka encoder. 

%This alignment ensures that our Matryoshka encoder is language agnostic and isomorphic between languages and that when comparing the semantic similarity, regardless of the languages used, the cosine similarity is similar.  

\noindent
\textbf{Hierarchical Clustering via Level-Wise Fine-Grained Similarity Calculations.} To cluster our Matryoshka embeddings for news article data, we propose a modification to Reciprocal Agglomerative Clustering 
(RAC)~\cite{sumengen2021scaling}. This is such that for the first round of merging reciprocal nearest neighbors (RNN)-- points that are both the most similar to each other -- in the RAC algorithm, we utilize the first $d/4$-dimensional representation for the top layer of $\ell=1$ of our hierarchy that represent the themes present in the dataset. These merges continue as long as the centroid similarities for any given clusters $\mu_i, \mu_j$ are above a given threshold $\lambda_1$. Subsequently, after all of the themes are generated using the $d/4$ dimensions, we again cluster the news stories within these clusters, creating the $\ell=2$ layer of our clustering, representing the topics in the dataset. We again combine the nearest neighbors until the maximum similarity between any two clusters is below a given threshold $\lambda_2$, except using the $d/2$-dimensional representation of the centroids from the first round (appropriately weighting for the number of data points within each cluster). To create the $\ell=3$ layer of the hierarchical clustering, we repeat this to construct the bottom layer of the tree using the full representation with another threshold $\lambda_3$ (see Appendix~\ref{sec:appendix-dp-stable-trees}). Throughout our work, we determine the $\lambda_{\ell}$ thresholds for combining RNNs empirically based on the value that achieves the highest $F_1$ score in differentiating news articles on the validation set from the SE-22-t8 dataset.

%\subsection{Training and Experiment Details}

%\subsection{}

%% file: results.tex
\section{Benchmarking and Results}
Having given an overview of our training scheme for our Matryoshka embeddings, we now benchmark their ability to differentiate news articles of varying similarity and evaluate the efficacy of our proposed clustering algorithm.

%To select the base contextual encoder model to which we apply our Matryoshka training methodology (see Methodology), as well as to develop several traditionally fine-tuned embeddings to which to compare our Matryoshka embedding
\subsection{Embeddings Evaluation}
%we additionally train traditional embeddings utilizing the original AngIE objective~\cite{li2023angle} to compare against.
%
As in past work, we evaluate our models using their Pearson's correlation $\rho$ with the  \textsc{overall} labels on the
test split of the SemEval 22 t8 dataset~\cite{chen2022semeval}. We evaluate our Matryoshka embedding approach against both fine-tuned and baseline versions of several popular multilingual encoder models with context windows of 512 tokens that were trained using the original unaltered AngIE objective function~\cite{li2023angle}. We additionally benchmark our models against the best bi-encoder model submitted to the SE-22-t8 task, \texttt{GateNLP-UShef}~\cite{singh2022gatenlp}.\footnote{We choose to include the numbers of our models against several pre-trained and non-fine-tuned open-source models given their widespread popularity in embedding news articles, tracking news events, and news topics~\cite{grootendorst2022bertopic,nielsen2022mumin,xu2022hfl}.}
%To benchmark our Matryoshka embeddings, we select popular  to serve as a base to then apply our modified AngIE objective function $\mathcal{L}_{AngIE}$.

\vspace{3pt}
\noindent
\textbf{Training Setup.}  While training our embeddings, we set the learning rate to $2\times 10^{-5}$ and use AdamW as the optimizer~\cite{kingma2014adam}. Due to computational constraints, while training on an NVIDIA A6000 GPU, we use a batch size of 16.  As in~\citet{li2023angle}, while training our Matryoshka embeddings utilizing the modified AngIE objective, we set the temperature parameter $\tau$ to 0.05. We use a maximum length of 512 tokens. We utilize 10\% (410K article pairs) of our SE-22-t8 dataset as validation. During training, we evaluate performance every 10K~training steps and use a patience of 2 for ending training. 

\vspace{2pt}
\noindent
\textbf{Evaluation Results.}
As seen in Table~\ref{tab:semeval-results}, our models trained with our modified AngIE objective achieve competitive scores for the former SE-22-t8 test set. Indeed, our models perform better than the previous state-of-the-art model, with our \texttt{fine-tuned-mE5-base} model achieving state-of-the-art results for bi-cross embedding models (embedding models can be utilized for clustering news articles) on the {test} split (0.817 vs.\ 0.801). 
\begin{table}[t]
    \centering
    \scriptsize
    \begin{tabular}{l|r|r}
    Model & \textbf{SE-22} & \textbf{SE-22 Ext.} \\ \midrule
    mE5-base & 0.604 & 0.582 \\% Good
    fine-mE5-base (ours)&\textbf{0.817} & {0.812} \\ %Good
    mpnet-base  & 0.513 & 0.520\\ %Good
    fine-mpnet-base (ours)& 0.791 &0.801\\ %Good
    xlm--roberta-base  & 0.225 & 0.119\\ %Good
    fine-xlm-roberta-base (ours) & 0.773 & 0.796\\ %Good
    umt5-base &0.262 & 0.119\\ %Good
    fine-umt5-base (ours) & 0.815 & 0.582\\ %Good

   mat-mE5-base-192 (ours) & 0.799  & 0.808\\
    mat-mE5-base-384 (ours) & 0.792& \textbf{0.816}\\
    mat-mE5-base-768 (ours) &0.676 &  0.776\\
   
    GateNLP-UShef & 0.801 & -- \\ \bottomrule
    \end{tabular}
    \vspace{-6pt}
    \caption{Comparison of different pre-trained bi-encoder models and our 
 fine-tuned models on the SE-22-t8 test dataset in terms of Pearson correlation.}
 %\vspace{-20pt}
 \vspace{-10pt}

    \label{tab:semeval-results}
\end{table}
Our embeddings further extend to the other added languages in our extended test dataset. Given the advantage of utilizing \texttt{mE5-base} in learning to differentiate between different news articles, we utilize the \texttt{mE5-base} encoder language model as our base model and focus on its Matryoshka-trained version for the remainder of our paper. This reinforces past work~\cite{park2024improving,wang2024multilingual} that suggests the mE5 model's success over prior models in modeling multilingual texts. See Appendix~\ref{sec:app-pearson} for additional bilingual results.

\begin{figure*}[!htbp]
 \centering
    \begin{subfigure}[b]{0.24\textwidth}
        \centering
        \includegraphics[width=\textwidth]{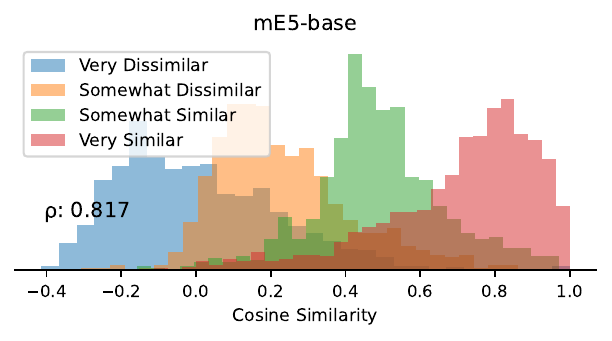}
        \caption{fine-mE5-base}
        \label{fig:d4-dimensions-1}
    \end{subfigure}
    \begin{subfigure}[b]{0.24\textwidth}
        \centering
        \includegraphics[width=\textwidth]{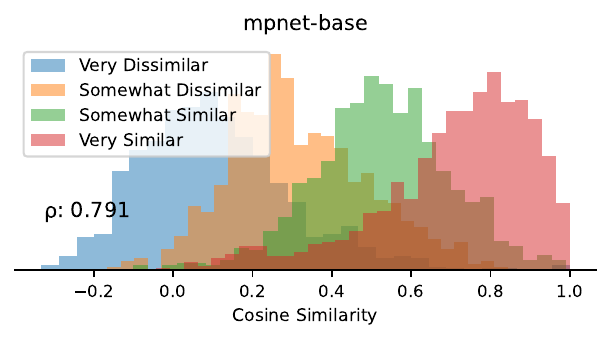}
        \caption{fine-mpnet}
        \label{fig:d4-dimensions-2}
    \end{subfigure}
    \hfill
    \begin{subfigure}[b]{0.24\textwidth}
        \centering
        \includegraphics[width=\textwidth]{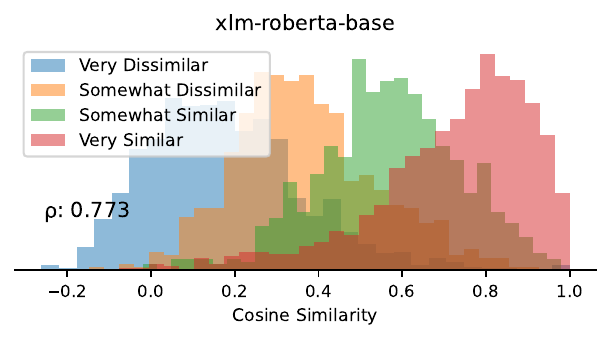}
        \caption{fine-xlm-roberta-base}
        \label{fig:d2-dimensions-3}
    \end{subfigure}
    \hfill
    \begin{subfigure}[b]{0.24\textwidth}
        \centering
        \includegraphics[width=\textwidth]{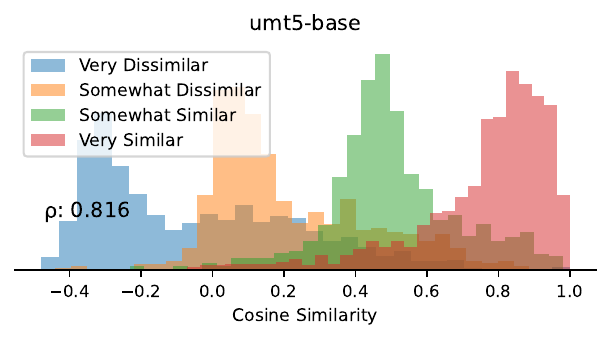}
        \caption{fine-umt5-base}
        \label{fig:d-dimensions-5}
    \end{subfigure}

    \centering
    \begin{subfigure}[b]{0.3\textwidth}
        \centering
        \includegraphics[width=\textwidth]{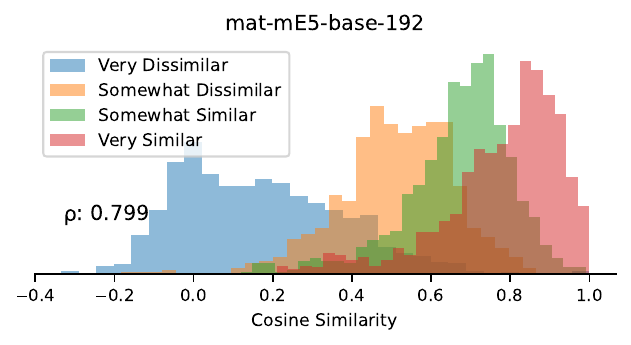}
        \caption{matryoshka-mE5-$d/4$ dimensions}
        \label{fig:d4-dimensions}
    \end{subfigure}
    \hfill
    \begin{subfigure}[b]{0.3\textwidth}
        \centering
        \includegraphics[width=\textwidth]{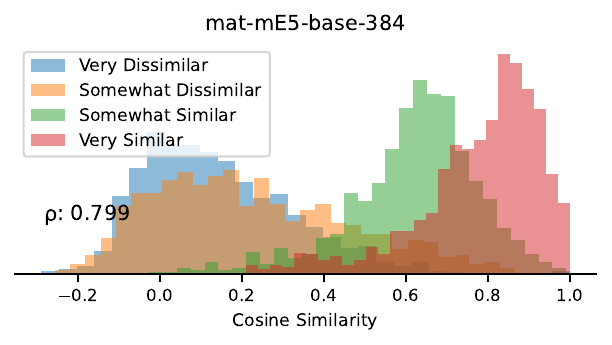}
        \caption{matryoshka-mE5-$d/2$ dimensions}
        \label{fig:d2-dimensions}
    \end{subfigure}
    \hfill
    \begin{subfigure}[b]{0.3\textwidth}
        \centering
        \includegraphics[width=\textwidth]{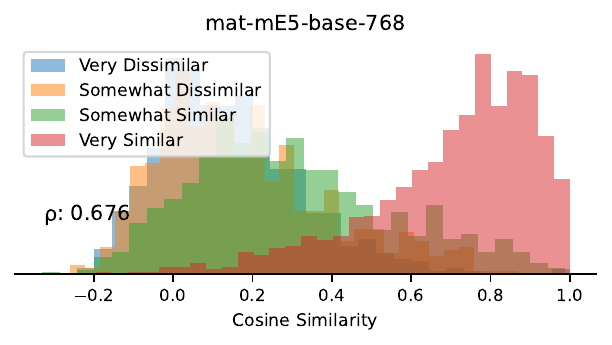}
        \caption{matryoshka-mE5-$d$ dimensions}
        \label{fig:d-dimensions}
    \end{subfigure}
    \caption{Cosine similarity of test split of the  SE-22-t8 dataset article pair embeddings utilizing Matryoshka e5-base embeddings. As more dimensions are added, the  Matryoshka embeddings can distinguish between greater article similarities.}
    \label{fig:dimensions}
    \vspace{-10pt}

\end{figure*}

Plotting the cosine similarities for the different similarity pairs within the test split of the SE-22-t8 dataset in Figure~\ref{fig:dimensions}, we observe a separation between the similarity pairs at different dimensions for our Matryoshka in comparison to the fine-tuned versions of our other embeddings, illustrating the ability of Matryoshka embeddings to learn increasingly detailed information about texts at higher dimensions. To further show this ability, we compute the AUROC for each of our models' abilities to differentiate news article similarity at various similarity thresholds based on their cosine similarity. As seen in Table~\ref{tab:differentiation-auroc}, across both the original test dataset and our extended 54-language test dataset, our Matryoshka embeddings largely perform the best at differentiating news articles at each similarity level compared to our traditionally trained embeddings.% This is the case for every similarity category. %As shown in Figure~\ref{fig:dimensions}, \texttt{umt-base} achieves sharp peaks in differentiating different types of news articles; however, it performs relatively poorly on our extended dataset (Table~\ref{tab:semeval-results}).

\subsection{Ablations and Relational Similarity}

\vspace{3pt}
\noindent
\textbf{Identical Positive Pairs with Dropout.} The inclusion of identical positive pairs with independently sampled dropout~\cite{gao2021simcse} is key to the success of our models. This approach forced our models to retain their language similarly to their similarities by ensuring slightly similar sentence embeddings in the \textit{same} language were pushed towards each other. We note that this further enabled us during training to artificially increase our batch size (\textit{i.e.}, more positives and negatives within each batch). Indeed, training a model based on the \texttt{mE5-base} encoder without these artificial in-batch negatives, our Pearson $\rho$ correlation on the SemEval 22-t8 dataset was $\rho =0.693$ using the first 192~dimensions, $\rho=0.702$ for the first 384~dimensions, and $\rho =0.684$ for the full embedding. We find similar behavior on the extended 54-language SemEval 22-t8 dataset with $\rho =0.733$ using the first 192~dimensions, $\rho=0.745$ for the first 384~dimensions, and $\rho =0.740$ for the full embedding. We further find that ridding our model of all in-batch negatives by removing $\mathcal{L}_{contrast}$ similarly 
hurts our model, with the Pearson correlation $\rho$ decreasing to nearly zero.
\vspace{3pt}
\noindent
\textbf{Data Augmentation.}
We also find that our data augmentation approach enabled our work to extend to more languages and led to better performance overall. To show this, we perform another ablation where we train our Matryoshka \texttt{mE5-base} encoder using only the SemEval 22-t8 dataset. While we do achieve high performance on the original SemEval test set $\rho =0.828$ using the first 192 dimensions, $\rho=0.835$ for the first 384 dimensions and $\rho =0.802$ for the full embedding, this lead to degraded performance on the extended 54-language test dataset ($\rho =0.706$ using the first 192 dimensions, $\rho=0.799$ for the first 384 dimensions and $\rho =0.712$).

\begin{table}
\centering
\scriptsize
\setlength{\tabcolsep}{4pt}  % Reduce the space between columns
\renewcommand{\arraystretch}{0.9}  % Reduce the space between rows
\begin{tabular}{lp{2.7cm}|ccc}  % Set a fixed width for the second column

{Dataset}  & Model & $\geq$ SD &$\geq$ SS & $\geq$VS \\ \toprule
SE-22-t8 & mat-mE5 (ours) & \textbf{0.948} &  \textbf{0.949} & \textbf{0.934}\\

& mE5-base & 0.850 &0.838 &0.799 \\
 & fine-mE5-base (ours)& 0.917 &0.939 &0.911 \\
 & mpnet-base  & 0.808 & 0.786& 0.732\\ %Good
 & fine-mpnet-base (ours)& 0.913 & 0.922 & 0.895 \\
 & xlm-roberta-base  & 0.693 & 0.678& 0.661\\ %Good
 & fine-xlm-roberta-base (ours) & 0.905 & 0.914 & 0.882\\
 & umt5-base  &0.715 & 0.704& 0.670 \\%Good
 & fine-umt5-base (ours)  & 0.902 & 0.935 &   0.926\\ \midrule
SE-22-t8-Ext. & mat-mE5 (ours) & \textbf{0.960} &  \textbf{0.967} & \textbf{0.962}\\

 & mE5-base & 0.890 & 0.879& 0.860\\
& fine-mE5-base (ours)& 0.953 & 0.962& 0.959 \\

 & mpnet-base & 0.847 & 0.838 & 0.813 \\
 & fine-mpnet-base (ours) & 0.952 & 0.961 & 0.949 \\
 & xlm-roberta-base  &0.626 & 0.617 &0.610  \\
 & fine-xlm-roberta-base  (ours) &0.949 & 0.960 & 0.948 \\
 & umt5-base &0.676 & 0.668&  0.652\\
 & fine-umt5-base (ours) &0.893 & 0.878 &  0.859\\
\bottomrule
\end{tabular}
\vspace{-6pt}
\caption{\label{tab:differentiation-auroc} AUROC for differentiating news article pair similarities on the test split of the SE-22-t8 dataset. }
\vspace{-10pt}
\end{table}

%For the Matryoshka-embeddings we progressively utilize more dimension (\textit{i.e.}, utilizing the first $d/2$ dimensions for distinguishing ``Somewhat Similar'' and ``Very Similar'' articles.)
%\su
%\
%We utilize the \texttt{mE5-base} as an initial checkpoint for training our Matryoshka embeddings and train for one epoch our extended SE-22-t8 dataset and subsequently perform multilingual alignment using the English translations of the dataset (see Methodology). 

\vspace{3pt}
\noindent
\textbf{Relational Similarity.} Finally, to ensure that our Matryoshka embeddings were isomorphic, we determined the relational similarity between the language's GP4o translated test set embeddings using our extended test SE-22 dataset. We observed an average of 0.753 relational similarity with English, with the lowest relational similarity being between English and Burmese at 0.452 (the highest being between English and Portuguese at 0.839). See Appendix~\ref{sec:isomorphism} for additional details. Based on prior work, we thus achieve an acceptable average multilingual alignment across our embeddings~\cite{conneau2017word,marchisio2022isovec}. We leave to future work to translate additional texts in low-resource languages like Burmese, Kannada, and Malayalam to potentially improve our relational similarity for these languages.

%After alignment, the average relational similarity was   X (significant difference via the Mann-Whitney U-test), with the lowest relational similarity being between Burmese at 0.456 and Kannada at 0.598 . We list all relational similarities pre and post-alignment in the Appendix. 

\subsection{Multilingual Clustering} Having observed that our models, in particular our Matryoshka model, are better able to differentiate news articles at varying levels of similarity, we now benchmark these embeddings using two common clustering techniques used for encoder embeddings: Mini-Batch K-Means\footnote{For Mini-Batch K-Means we specify the number of clusters and the batch size as 32.} and BERTopic~\cite{grootendorst2022bertopic}. BERTopic is a recent, widely used technique that utilizes the UMAP~\cite{mcinnes2018umap} and the HDBSCAN density-based clustering algorithm~\cite{mcinnes2017hdbscan} to group together documents from their embeddings. For both of these algorithms, we determine their performance on~\cite{miranda2018multilingual}'s multilingual news article dataset and the 20 Group News dataset, which groups together documents by general theme. As in~\citet{miranda2018multilingual}, we benchmark our embeddings by computing our embeddings' $F_1$ scores for correctly grouping documents using BERTopic and mini-batch K-Means.
\begin{table*}
\scriptsize
\centering
\begin{tabular}{l|l|ccc|ccc}
\toprule
%\scriptsize
\textbf{} & \textbf{} & \multicolumn{3}{c|}{\textbf{BERTopic}} & \multicolumn{3}{c}{\textbf{MiniBatch-KMeans}}  \\ \hline
\textbf{Data}  & \textbf{Model} &  \textbf{Prec.} & \textbf{Recall} & \textbf{$F_1$} & \textbf{Prec.} & \textbf{Recall} & \textbf{$F_1$} \\ \hline
\citet{miranda2018multilingual} & xlm-roberta-base & 0.3568 & 0.3641 &  0.3604 & 0.5171 & 0.0964 & 0.1626 \\
& fine-xlm-roberta  & 0.6810 & 0.8116& 0.7406&     0.6432 & 0.1293 &  0.2153  \\
 & mpnet-base & 0.8388 & 0.6830 &  0.7529 & 0.6676 & 0.1424 & 0.2348 \\
 & fine-mpnet-base&0.7071 & {0.8415}&{0.7685} &   0.6600 & 0.1305 &  0.2179 \\
 & umt5-base & 0.0665 & 0.2960 & 0.1086 &0.1536 & 0.0638 & 0.0902 \\
 & fine-umt5-base &0.4983 & 0.5360& 0.5165 & 0.5692 & 0.0864 &  0.1501  \\
 & mE5-base & \textbf{0.8507} & 0.3715 & 0.5171 & 0.4919 & \textbf{0.1960} & \textbf{ 0.2803}\\
 & fine-mE5-base &0.7791 & 0.5735 & 0.6607  & \textbf{0.7111}&  0.1468 & 0.2434 \\
 & mat-mE5-base-192 & {0.7895}& \textbf{0.8971}&\textbf{0.8399} & {0.6876} & 0.1271 & 0.2145\\

 & mat-mE5-base-384 &0.7798&0.8496 &{0.8132} & 0.6761 & 0.1232 & {0.2083}\\
  & mat-mE5-base-768 & 0.7661& {0.8677}& 0.8137&0.6329 &0.1338&  0.2209 \\
 \midrule
 20 NewsGroup & xlm-roberta-base & 0.0504 & \textbf{0.9372} & 0.0956 & 0.1004 & 0.1297 & 0.1132 \\
& fine-xlm-roberta & 0.0967 & 0.6783 & 0.1693 &0.1736 & 0.4480 &  0.2502\\
 & mpnet-base & 0.0692 & 0.8415 & 0.1280 & 0.2086 & 0.5216 & 0.2980 \\
 & fine-mpnet-base & 0.1193&0.7057 &0.2041 & 0.2009&0.4933 &  0.2855 \\
 & umt5-base & 0.0504 & 0.9374 & 0.0956 & 0.0617 & 0.1271 & 0.0831 \\
 & fine-umt5-base & 0.0505& 0.8737&0.0955 &  0.1284 & 0.1688 & 0.1459   \\
 & mE5-base & 0.0644 & 0.8755 & 0.1199 & 0.1550 & \textbf{0.5724} & 0.2439 \\
 & fine-mE5-base  &\textbf{0.1813}&0.5589 & \textbf{0.2738}&{0.2258} & 0.4775 & 0.3066 \\

  & mat-mE5-base-192 &0.1711 &{0.6743} &  0.2730& 0.2131 &  0.4642 & 0.2921  \\

 & mat-mE5-base-384  & {0.1622}&0.6027 &{0.2606} &\textbf{0.2131} & 0.5236 &  \textbf{0.3044}  \\
  & mat-mE5-base-768 &  0.1521& 0.2818 & 0.1976 & 0.0947 &0.2310 & 0.1343\\
 \bottomrule
\end{tabular}
\vspace{-8pt}
\caption{\label{tab:differentiation-clustering} Performance for clustering news articles on the Miranda and 20 NewsGroup datasets.  }
\vspace{-10pt}

\end{table*}

As seen in Table~\ref{tab:differentiation-clustering}, our models, particularly our Matryoshka model, perform largely the best in terms of differentiating news articles, both utilizing BERTopic and Mini-Batch K-Means.  This is particularly true when utilizing BERTopic, where the Matryoshka model utilizing the first 192 dimensions achieves an $F_1$ score of 0.8399 on the~\citet{miranda2018multilingual} dataset and an $F_1$ score of 0.2730 on the 20 NewsGroup dataset using BERTopic. We note that our fine-tuned models largely outperform their generic counterparts.

\vspace{3pt}
\noindent
\textbf{Hierarchical Clustering.}
We additionally benchmark our approach for hierarchical clustering. Given the lack of hierarchical-level text clustering benchmarks to determine the performance of our hierarchical clustering algorithms we  (1) cluster the {test} split of SE-22-t8 dataset at various similarity granularities utilizing our proposed hierarchical clustering algorithm as well as BERTopic and (2) subsequently again determine the $F_1$ scores of these clustering algorithms in correcting grouping ``Very Similar'' articles together at the highest granularity, ``Somewhat Similar'' and higher articles the second highest granularity, and finally ``Somewhat Dissimilar'' news articles at the lowest granularity. We test our best-performing models, the Matryoshka embedding and our traditionally trained mE5-base and mpnet embeddings, on this task utilizing our RAC algorithm and BERTopic~\cite{grootendorst2022bertopic}. For our RAC algorithm, we utilize the optimal $\lambda_{\ell}$ for differentiating news article similarities determined from the held-out validation set of the SE-22-t8 dataset. %For the original RAC algorithm we progressively merge clusters at these thresholds  

\begin{table}[t]
    \centering
    \scriptsize
    \begin{tabular}{l|ccc}
   \toprule
   %\scriptsize
   \textbf{fine-mpnet-base} & SD & SS & VS \\ \hline
    BERTopic &0.817 & 0.693 & 0.616\\  \hline
     Level wise RAC  & 0.822 & 0.775 &0.724 \\ \midrule
    \textbf{fine-mE5-base} & SD & SS & VS \\ \hline
   
    BERTopic &0.802 & 0.648 & 0.579\\  \hline
     Level wise RAC  & 0.822 & 0.775 &0.752 \\ \midrule
    \multicolumn{1}{l}{\textbf{mat-mE5}} & SD & SS & VS   \\  \hline
    BERTopic & 0.819 & 0.738 & 0.608 \\ \hline
    Level wise RAC & \textbf{0.849} & \textbf{0.816} &\textbf{0.795}  \\ \bottomrule
    \end{tabular}
    \vspace{-6pt}
    \caption{Comparison of the $F_1$ score in clustering documents in the SE-22-t8 dataset at various levels of granularity using our approach versus BERTopic.}
    \label{tab:clustering-results}
    \vspace{-15pt}
\end{table}

As seen in Table~\ref{tab:clustering-results}, our level-wise RAC approach outperforms BERTopic in differentiating the SE-22-t8 dataset at every level of granularity. Similarly, our Matryoshka mE5 model outperforms our traditionally trained AngIE mE5 and mpnet models using our level-wise RAC approach.

%best across our SE-22-t8, and across both the datasets our DP Tree algorithm performs better than Stable Greedy Trees. We see regardless of the dataset that our approach largely succeeds over just a baseline fine-tuned approach or utilizing pre-trained embeddings that we can identify individual news stories with better reliability. Finally, to ensure that our results generalize beyond the SE-22-t8 dataset, we cluster the test split of the Wikipedia Current Portal dataset~\cite{gholipour-ghalandari-etal-2020-large}, which consists of 1,022 news events and a corresponding set of 70,839 news articles about those events. We embed these news articles utilizing each of our models at the highest level of granularity and subsequently determine the percentage of the news article pairs that were about the same event that were clustered in the same cluster. Given the success of the RNN algorithms as seen in Table~\ref{tab:clustering-results}, we focus on this algorithm for this experiment. In this experiment, for the \texttt{gte-multilingual-base} embeddings, we attain a score of 0.523, for the \texttt{finetuned-mE5-base} a score of 0.567, and for \texttt{Matryoshka-mE5} a score of 0.624.

\noindent
\textbf{Case Study: Interpretable Multilingual Clusters.}
While clustering multilingual news articles can identify different news stories amongst international news websites and social media data, these clusters can still be difficult to parse and understand. To create human-interpretable representations of the underlying clusters, we take two approaches: (1) English story-level summaries, and (2) representative English keyword extraction at each level of the hierarchical clustering. 

For our story-level summaries, we fine-tune a LLaMA model\footnote{\url{huggingface.co/meta-llama/Llama-3.1-8B-Instruct}} to perform multilingual news article summarization and output English-language summaries~\cite{dubey2024llama}. Specifically, to train this model, we first translate documents for each language from the multiple-document summarization dataset Multi-News~\cite{fabbri2019multi}. We subsequently construct a dataset of 38,830 multilingual training examples, 7,726 validation examples, and 7,521 test examples, where individual training instances are made up of multiple documents in a variety of one of our 54 different languages. We subsequently fine-tune our LLaMa model to output English summaries from these multilingual multi-document examples (see Appendix~\ref{sec:finetune-lllama}).

For our representative English keywords, for each identified news story cluster, we first extract keywords from each of these summaries using the class-based TF-IDF method proposed by~\cite{grootendorst2022bertopic}. This method essentially treats each English summary of each news story-level cluster as one document and performs traditional TF-IDF keyword extraction (we utilize this method rather than looking directly at the individual documents, given the multilingual setting)~\cite{ramos2003using}. Upon extracting these keywords, we then extract keywords for the next level of clusters (the topics), by again performing class-based TF-IDF on the larger clusters that are groupings of the summaries of the lower fine-grained clusters. This is such that we treat the summaries of the news-story clusters that form the large topic-level cluster as one document, and again perform TF-IDF. We utilize the same method for the top-level theme clusters.

\begin{table}
\centering
\scriptsize
\begin{tabular}{lr}
\toprule
 \textbf{Story Keywords} & \textbf{Articles}\\ \hline
emails, sony, interview, circulate, dumping, bitter & 872 \\
choice, announce, senate, decision, attorney, inquiring & 669\\
aviation, icao, commercial, black, plane, global & 407\\
celtics, turnovers, season, raptors, loss, outrebounded & 359\\
girl, baby, hospital, prince, kate, little & 294\\ \midrule

 \textbf{Topic Keywords} & \textbf{Articles}\\ \hline
isis, obama, syria, said, state & 2,311 \\
fifa, blatter, officials, corruption, soccer & 1,953\\
messi, team, chelsea, coach, club & 1,273 \\
sony, movie, theaters, pictures, hackers & 1,035 \\
plane, pilot, flight, crashed, cockpit & 972  \\\midrule

 \textbf{Theme Keywords} & \textbf{Articles}\\ \hline
fifa, blatter, officials, world, corruption & 4,792 \\
obama, president, said, state, isis & 2,643 \\
sony, apple, car, movie, film & 1,903\\
plane, pilot, flight, crashed, airport & 976\\
police, man, navalny, nisman, shot & 796\\
\bottomrule
\end{tabular}
    \vspace{-6pt}
\caption{\label{tab:narratives-miranda}~\citet{miranda2018multilingual} --- Top stories, topics and themes.} %\todo{how was this selection made? Sounds vague.
\vspace{-10pt}
\end{table}

\begin{table}
\centering
\scriptsize
\begin{tabular}{lr}
\toprule
 \textbf{Story Keywords} & \textbf{\# Articles}\\ \midrule
rcn, strikes, ballot, members, chink & 139\\
chapman, intent, murder, christmas, elle & 97\\
sharp, loan, conflicts, chairman, potential & 96\\
contest, acerbic, lydon, clown, nora & 90\\
lemur, zoo, species, dancing, coquerel & 88 \\  \midrule
 \textbf{Topic Keywords} & \textbf{\# Articles}\\ \hline
police, old, murder, man, year & 1,762 \\
says, uk, police, government, company & 1,748\\
israel, gaza, hamas, military, palestinian, & 819\\
ukraine, russia, zelensky, president& 710\\
people, boat, coast, turkey, city &  565\\
\midrule

 \textbf{Theme Keywords} & \textbf{\# Articles}\\ \hline
says, uk, government, new, police & 3,339\\
police, old, man, year, murder & 3,136\\
ukraine, russia, russian, ukrainian, gaza & 1,838\\
party, government, labour, election, minister & 1,383\\
team, world, cup, match, league & 1,178\\
\bottomrule
\end{tabular}
\vspace{-6pt}
\caption{\label{tab:narratives-bbc} BBC---Top stories, topics and themes.} %\todo{how was this selection made? Sounds vague.
\vspace{-12pt}
\end{table}

\begin{table}[h]
\centering
\scriptsize
\begin{tabular}{p{0.7\linewidth}r} % p for wrapped text and r for right alignment
\toprule
\textbf{Top Story} & \textbf{Articles} \\ 
\midrule
\raggedright \textbf{Miranda Story:}{ The hackers who have been dumping Sony\'s emails and documents online have now threatened violence against theaters showing the upcoming Seth Rogen and James Franco movie The Interview, which is about two Americans who plot to assassinate North Korean leader Kim Jong Un. The threat, written in broken English, was posted on file-sharing services that have been used to circulate internal Sony emails stolen in the cyberattack, reports Variety.} & 872 \\

\midrule
\raggedright \textbf{BBC Story:} {The Royal College of Nursing (RCN) has announced it will ballot its members for a further strike, with the union\'s general secretary Pat Cullen saying there is a "chink of optimism" ahead of talks with the government. The RCN has not ruled out further strikes but says it will "continue to engage in constructive dialogue" with the government. The RCN has been in dispute over pay since last year and has held two strikes since the start of January.  }& 139\\

\bottomrule
\end{tabular}
\vspace{-6pt}
\caption{\label{tab:narratives-top-stories} Top news story in each dataset.}
\vspace{-6pt}
\end{table}
%As in~\citet{sarthiraptor}, we propose utilizing recursive summaries for summaries higher in our the tree produced by our level-wise RAC approach (\textit{i.e.} somewhat similar summaries are constructed as summaries of the very similar summary clusters, \textit{etc...}).

To illustrate the use of our work in tracking news stories, we cluster all the BBC articles from 2023\footnote{\url{https://huggingface.co/datasets/RealTimeData/bbc_news_alltime}} and the~\citet{miranda2018multilingual} dataset using our Matryoshka embeddings and the Level-Wise RAC algorithm. We list the top news stories and their corresponding number of articles in Tables~\ref{tab:narratives-miranda} and~\ref{tab:narratives-bbc}  and the top story summaries in Table~\ref{tab:narratives-top-stories}. As qualitatively seen, our method can effectively extract different levels of granularity meaning in the form of individual stories, topics, and themes.

\section{Conclusion}
In this work, we introduce a novel application of multilingual Matryoshka embeddings that leverages their hierarchical structure to distinguish news articles at varying levels of granularity. Building on this, we propose a hierarchical clustering approach based on the Reciprocal Agglomerative Clustering algorithm to effectively identify individual news stories, broader topics, and overarching themes within news datasets.  

%We develop two novel hierarchical algorithms, illustrating how their application in concert with our Matryoshka embeddings can lead to better clustering of news articles. Our approach will allow for the clustering and identification of news articles and themes across languages, thus enabling an open-source means of understanding the globalized news ecosystem.

\section*{Limitations}
We note the limitations of our approach here.

\noindent
\textbf{Use of GPT-4o to Extend our Dataset.} While synthetic data has been shown to improve performance across various datasets ~\cite{gandhi2024better,hurst2024gpt}, as shown by~\citet{li2023synthetic}, training solely on synthetic data can lead to subpar performance. However, as noted~\cite{li2023synthetic}, this is largely true for subjective tasks, and this can be mitigated by guiding large language models with real-world examples. For this reason, throughout this work, we mostly utilize GPT-4o for translations, depending largely on real-world articles rather than having GPT-4o construct articles for our dataset.

\noindent\textbf{Small Batch Sizes.} Due to hardware limitations (Nvidia RTX A6000), we train our models with a batch size of 16. As noted elsewhere~\cite{chen2022we}, large batch sizes are typically needed when performing contrastive learning; as such, some of our results could largely be improved utilizing a larger batch size. %We leave to future work to train our models utilizing larger batch sizes for higher performance.

\section*{Ethics Statement}
For scraping our dataset, we largely rely on the code provided by~\citet{chen2022semeval}. While scraping, we limit the load that each news site experiences by scraping at a maximum rate of one
request every 10 seconds. We further note that the
hosts that we scrape from are identifiable through WHOIS,
reverse DNS, and an HTTP landing page. This page explains how to
reach us if the website wishes to opt-out of scraping. 
We received no requests from websites to opt out.

To respect the original authors of each news article, we release only the synthetic portions of our dataset. While we do not see any additional risks from releasing our work, we note that it could be utilized in works like~\citet{hanley2024specious} to understand the spread of narratives amongst news websites and potentially target websites that are influential in spreading specific narratives. 
%Given that our dataset is partly made of translations of the \texttt{VAST}, \texttt{C-STANCE}, and \texttt{EZ-STANCE} datasets as well as data constructed synthetically from GPT-4o, we make our dataset publically available. We lastly note that our work, while improving upon multilingual stance detection and providing a new dataset from training and testing multilingual stance detection models, our model does not perform evenly across different languages (performing the worst on languages like Burmese and Kannada). We leave to future work to improve our model's ability to perform well across under-resourced languages~\cite{kumar2024negatived}.

%% file: appendix.tex
\newpage
\section{News Paper Article Pairs at Each Level of Similarity\label{sec:pair-examples}}
\begin{figure}[!h]
\begin{minipage}{.47\textwidth}
\centering \small
\textbf{Very Similar}
\end{minipage}
\noindent\fcolorbox{black}{lightgray}{%
\begin{minipage}{.47\textwidth}
\small
\textbf{Article 1:} My Dear Kogites, It gives me great joy to welcome everyone to this New Year and I thank God Almighty for keeping us alive and healthy to witness it. 2020 is also the beginning of a new decade commencing today, being Wednesday, the 1st of January, 2020 and ending on Monday, the 31st day of December, 2029, making it a compelling point to set long term personal and state development goals.

\textbf{Article 2:} A former gubernatorial aspirant on the platform of All Progressives Congress (APC) in Kogi State, Gen. Patrick Ademu Akpa has urged Nigerians to be positive about the new year.
\end{minipage}}
\vspace{2pt}

\begin{minipage}{.47\textwidth}
\centering \small
\textbf{Somewhat Similar}
\end{minipage}
\noindent\fcolorbox{black}{lightgray}{%
\begin{minipage}{.47\textwidth}
\small
\textbf{Article 1:} Lady Gaga got a midnight kiss from a mystery man.The "Shallow" singer, 33, was spotted kissing a man who wasn't Dan Horton on New Year's Eve in Las Vegas following her residency performance. Wearing a sequined gown, Gaga and her new beau passionately made out as confetti fell around them.

\textbf{Article 2:}LADY GAGA 'DEVASTATED' AS SHE PULLS PLUG ON LAS VEGAS GIG AT LAST MINUTE. The pop star and actress celebrated the New Year in style following her recent split from fiance Christian Carino. Lady Gaga snogs mystery man as she rings in New Year as a single lady.
\end{minipage}}
\vspace{2pt}

\begin{minipage}{.47\textwidth}
\centering \small
\textbf{Somewhat Dissimilar}
\end{minipage}
\noindent\fcolorbox{black}{lightgray}{%
\begin{minipage}{.47\textwidth}
\small
\textbf{Article 1:} Shares of CBRE Group (CBRE) have been strong performers lately, with the stock up 8.8\% over the past month. The stock hit a new 52-week high of \$61.54 in the previous session. CBRE Group has gained 53.1\% since the start of the year compared to the 0\% move for the Zacks Finance sector and the 0\% return for the Zacks Real Estate - Operations industry.

\textbf{Article 2:} Investors seek growth stocks to capitalize on above-average growth in financials that help these securities grab the market's attention and produce exceptional returns. But finding a great growth stock is not easy at all. By their very nature, these stocks carry above-average risk and volatility. 
\end{minipage}}
\vspace{2pt}

\begin{minipage}{.47\textwidth}
\centering \small
\textbf{Very Dissimilar}
\end{minipage}
\noindent\fcolorbox{black}{lightgray}{%
\begin{minipage}{.47\textwidth}
\small
\textbf{Article 1:} Telepathy is the ability to transmit words, emotions, or images to someone else's mind. Telekinesis is the ability to move objects through mind power. Teleportation refers to transporting yourself or your mind to a location miles away from you infraction of seconds.

\textbf{Article 2:} Every change begins with a leadership decision. Making the decision to institute changes is not always easy. Being prepared, planning well, and being surrounded by a good team will make that decision a lot easier. Begin by putting yourself in a positive frame of mind. 
\end{minipage}}
\label{figure:all-paragraph_pairs1}
\end{figure}

%\clearpage

\section{Example GPT-4o rewritten Articles\label{sec:appendix-rewritten}}

\begin{figure}[!h]
\noindent\fcolorbox{black}{lightgray}{%
\begin{minipage}{.5\textwidth}
\small

\textbf{Original Article:} Jewish groups have expressed shock and horror over the stabbings that occurred at a rabbi's house in the New York City suburb of Monsey, N.Y., on the evening of Dec. 28.The attack occurred at Chabad of Suffern Rabbi Chaim Rottenberg's home as he was lighting the candles on a menorah to celebrate the seventh night of Hanukkah. The attacker stabbed five people, including Rottenberg's son.

\textbf{Rewritten Article:} Jewish Communities Devastated by Monsey Stabbings during Hanukkah Celebration. In a harrowing incident on the evening of December 28, a stabbing attack at Rabbi Chaim Rottenberg's residence in Monsey, New York, left the Jewish community in shock. The brutal assault occurred as Rabbi Rottenberg was lighting the menorah candles for the seventh night of Hanukkah. Among the five victims, including Rottenberg's own son, two remained in critical condition as of December 30, while the other three were discharged from the hospital.

\textbf{Rewritten Article:} In an alarming incident that has left the Jewish community in shock, an attacker stabbed five people at a rabbi's residence in Monsey, New York, on December 28. The horrifying event took place during the lighting of Hanukkah candles at Rabbi Chaim Rottenberg's home. Among the injured was Rabbi Rottenberg's son. As of December 30, two of the victims remained in critical condition, while the remaining three were released from the hospital.

\textbf{Rewritten Article:} Devastating Monsey Stabbings Stir Outcry Among Jewish Communities. A state of shock and horror rippled through Jewish communities after a violent stabbing attack at a rabbi's residence in Monsey, NY, during Hanukkah festivities. On the night of Dec. 28, amidst the lighting of the menorah at Chabad of Suffern Rabbi Chaim Rottenberg's home, an assailant brutally stabbed five individuals, including the rabbi's son.

\end{minipage}}

\noindent\fcolorbox{black}
{lightgray}{%
\begin{minipage}{.5\textwidth}
\small
\textbf{Original Article:} The government on Wednesday slashed import duty on refined palmolein from 50 per cent to 45 per cent, while that on crude palm oil (CPO) from 40 per cent to 37.5 per cent with immediate effect, a move that the industry opposed saying it will hurt domestic refiners. A notification in this regard has been issued by the finance ministry. The duty cut has been made under the ASEAN agreement and the India-Malaysia Comprehensive Economic Cooperation Agreement (IMCECA), the notification added.

\textbf{Rewritten Article:}Import Duty Cuts Spark Controversy in Palm Oil Industry. In a significant policy shift, the government has reduced import duty on refined palmolein from 50\% to 45\%, and on crude palm oil (CPO) from 40\% to 37.5\%, according to a notification from the finance ministry issued Wednesday. This adjustment, rooted in the ASEAN agreement and the India-Malaysia Comprehensive Economic Cooperation Agreement (IMCECA), has met with strong opposition from domestic palm oil refiners on and intriguing novelty. This adjustment, rooted in the ASEAN agreement and the India-Malaysia Comprehensive Economic Cooperation Agreement (IMCECA), has met with strong opposition from domestic palm oil refiners.

\textbf{Rewritten Article:} Government Reduces Import Duties on Palm Oil, Sparks Industry Backlash. In a significant policy shift, the government announced on Wednesday a reduction in import duties on refined palmolein and crude palm oil (CPO), effective immediately. The duty on refined palmolein has been cut from 50\% to 45\%, and on CPO from 40\% to 37.5\%. This decision, issued by the Ministry of Finance, falls under the ASEAN agreement and the India-Malaysia Comprehensive Economic Cooperation Agreement (IMCECA).

\textbf{Rewritten Article:} Government's Import Duty Cut on Palm Oil Sparks Industry Concern. Date: Wednesday. In a move that has stirred widespread concern within the domestic refining industry, the government has announced an immediate reduction in import duties on palm oil. The duty on refined palmolein has been slashed from 50\% to 45\%, while crude palm oil (CPO) will see a reduction from 40\% to 37.5\%. This decision, formalized through a notification issued by the finance ministry, falls under the purview of the ASEAN agreement and the India-Malaysia Comprehensive Economic Cooperation Agreement (IMCECA).
\end{minipage}}

\label{figure:rewritten-pairs}
\end{figure}

\clearpage
\newpage
\section{Languages Utilized\label{sec:languages}}
We restrict our model to a subset of languages that OpenAI models support.\footnote{\url{https://help.openai.com/en/articles/8357869-how-to-change-your-language-setting-in-chatgpt\#h_513834920e}} These include the following: 
Albanian,
Arabic,
Armenian,
Bulgarian,
Burmese,
Catalan,
Simplified Chinese,
Czech,
Danish,
Dutch,
English,
Estonian,
Finnish,
French,
Georgian,
German,
Greek,
Gujarati,
Hindi,
Hungarian,
Icelandic,
Indonesian,
Italian,
Japanese,
Kannada,
Kazakh,
Korean,
Latvian,
Lithuanian,
Macedonian,
Malay,
Malayalam,
Marathi,
Mongolian,
Norwegian,
Persian,
Polish,
Portuguese,
Punjabi,
Romanian,
Russian,
Serbian,
Slovak,
Slovenian,
Somali,
Spanish,
Swedish,
Tamil,
Telugu,
Thai,
Turkish,
Ukrainian,
Urdu,
Vietnamese,

\section{Angle Objective\label{sec:app-angle-obj}}

As discussed by Li~et~al.~\cite{li2023angle}, while using contrastive learning has resulted in high-quality embeddings across semantic similarity tasks~\cite{gao2021simcse}, a common issue in training embeddings using contrastive learning is vanishing gradients due to their reliance on cosine similarity. By incorporating angle optimization in the complex space between embeddings, \citet{li2023angle} achieves improved results for embeddings on semantic similarity test datasets.

As in~\cite{sunrotate} and~\cite{li2023align}, we obtain the complex representations of contextual embeddings using the chunking strategy Namely, given the contextual representations of  a text pair $(\mathbf{h}_i \mathbf{h}_j)$, as originally calculated by Sun~et~al.~\cite{sunrotate, li2023angle}, their representations in the complex space are defined as follows:

\begin{equation}
z_{\mathbf{h}_i} = a + bi \in \mathbb{C} \quad \text{and} \quad w_{\mathbf{h}_j} = c + di \in \mathbb{C},
\end{equation}

where $a = \mathbf{h}_i^{re} \in \mathbb{R}$, $b = \mathbf{h}_i^{im} \in \mathbb{R}$, $c = \mathbf{h}_j^{re} \in \mathbb{R}$, and $d = \mathbf{h}_j^{im} \in \mathbb{R}$.  The angle between the complex representations of the 

The angle difference between $z_{\mathbf{h}_i}$ and $w_{\mathbf{h}_j}$, is calculated using polar coordinates as follows:

\begin{equation}
\frac{z_{\mathbf{h}_i}}{w_{\mathbf{h}_j}} = \gamma e^{i \Delta \theta_{zw}}
\end{equation}

\[
\gamma = \frac{|z_{\mathbf{h}_i}|}{|w_{\mathbf{h}_j}|} = \frac{\sqrt{a^2 + b^2}}{\sqrt{c^2 + d^2}},
\]

\[
\Delta \theta_{zw} = \theta_z - \theta_w,
\]
where $\theta_z$ and $\theta_w$ denote the respective angles of $z_{\mathbf{h}_i}$ and $w_{\mathbf{h}_j}$. The value of of $\frac{z_{\mathbf{h}_i}}{w_{\mathbf{h}_j}}$ is further computed as 

\begin{equation}
\frac{z_{\mathbf{h}_i}}{w_{\mathbf{h}_j}} = \frac{a + bi}{c + di} = \frac{(ac + bd) + (bc - ad)i}{c^2 + d^2}.
\end{equation}

Finally, the difference in the angle between $z_{\mathbf{h}_i}$ and  $w_{\mathbf{h}_j}$ is: 

\begin{align}
\Delta \theta_{zw} &= \text{abs}\left(\frac{z}{w} \times \frac{1}{\gamma}\right) \notag \\
&= \text{abs} \left[ \frac{(ac + bd) + (bc - ad)i}{\sqrt{(c^2 + d^2)(a^2 + b^2)}} \right].
\end{align}

%\newpage
\section{Hierarchical RNN  Algorithm\label{sec:appendix-dp-stable-trees}}
\begin{algorithm}[!h]
\caption{Level-Wise RAC}
\begin{algorithmic}[1]
\State \textbf{Input:} $\boldsymbol {\lambda}$,  $\mathbf{X} = \{x_i\}_{i=1}^N \in \mathbb{R}^d$ 
 \For{$m\in \{d/4, d/2,d \}$}
\If{$m =d/4$}
    \State $\phi_k^{{1:m}} \gets x^{1:m} $
\EndIf
\While { max(Cluster Similarities) $> \lambda_m $} \Comment{run the RAC algorithm}
\State Find Reciprocal Nearest Neighbors($\phi$)
\State Update Cluster Similarities($\phi'$)
\State Update Nearest Neighbors($\phi'$) \} 
%\State \Return RAC($C', \mathcal{M}$)
\EndWhile

 \State $n_k^m \gets |\{i : x_i^m = k\}|$

        \State $\phi_k^{{1:2m}} \gets \frac{\sum_{i : x_i = k} x^{1:2m} }{n_k}$
\EndFor
\end{algorithmic}
\end{algorithm}
\clearpage
\newpage
\section{Cross-lingual Isomorphism Metric\label{sec:isomorphism}} Within this work, we determine the cross-lingual isomorphism between English and the 53 other languages within our dataset to ensure that our model is language agnostic. We utilize the relational similarity metric that determines the correlation between the pairwise examples in different languages. 

\textbf{Relational Similarity}
Given seed translation pairs $(x_0, y_0), (x_1, y_1)$, the relational similarity is computed as follows by calculating the Pearson correlation in the following list~\cite{marchisio2022isovec,zhang2019girls}: 

\[
\begin{matrix}
\cos(x_0, x_1) & \cos(y_0, y_1) \\
\cos(x_0, x_2) & \cos(y_0, y_2) \\
\cos(x_0, x_3) & \cos(y_0, y_3) \\
\vdots & \vdots \\
\cos(x_1, x_0) & \cos(y_1, y_0) \\
\cos(x_1, x_2) & \cos(y_1, y_2) \\
\vdots & \vdots \\
\cos(x_s, x_t) & \cos(y_s, y_t)
\end{matrix}
\]

\begin{table}[t]
    \centering
    \small
    \begin{tabular}{l|r}
    \textbf{Language Pair}  & Relational Similarity \\ \hline
    English-Albanian & 0.786  \\
English-Arabic & 0.738  \\
English-Armenian & 0.704  \\
English-Bulgarian & 0.776 \\
English-Burmese & 0.452  \\
English-Catalan & 0.796  \\
English-Chinese & 0.743  \\
English-Czech & 0.810  \\
English-Danish & 0.831  \\
English-Dutch & 0.825  \\
English-English & 1.000  \\
English-Estonian & 0.770  \\
English-Finnish & 0.775  \\
English-French & 0.819  \\
English-Georgian & 0.709  \\
English-German & 0.823  \\
English-Greek & 0.737  \\
English-Gujarati & 0.631  \\
English-Hindi & 0.751  \\
English-Hungarian & 0.783  \\
English-Icelandic & 0.756  \\
English-Indonesian & 0.814  \\
English-Italian & 0.828  \\
English-Japanese & 0.707  \\
English-Kannada & 0.608  \\
English-Kazakh & 0.740  \\
English-Korean & 0.712  \\
English-Latvian & 0.765  \\
English-Lithuanian & 0.769  \\
English-Macedonian & 0.756  \\
English-Malay & 0.819  \\
English-Malayalam & 0.607  \\
English-Marathi & 0.706  \\
English-Mongolian & 0.710 \\
English-Norwegian & 0.829  \\
English-Persian & 0.731  \\
English-Polish & 0.803  \\
English-Portuguese & 0.839  \\
English-Punjabi & 0.639  \\
English-Romanian & 0.824  \\
English-Russian & 0.799  \\
English-Serbian & 0.797  \\
English-Slovak & 0.788  \\
English-Slovenian & 0.788  \\
English-Somali & 0.725  \\
English-Spanish & 0.802  \\
English-Swedish & 0.840  \\
English-Tamil & 0.650  \\
English-Telugu & 0.643  \\
English-Thai & 0.755  \\
English-Turkish & 0.794  \\
English-Ukrainian & 0.775  \\
English-Urdu & 0.725  \\
English-Vietnamese & 0.797  \\
    \end{tabular}
    \caption{Relational similarity scores for each of our languages. }
    \label{tab:alignment-results}
\end{table}
\clearpage
\onecolumn
\newpage

\newpage
\twocolumn
\section{Finetuning LLaMa-3.1 for Multilingual Multi-document Summarization\label{sec:finetune-lllama}}
We fine-tune a \texttt{LLaMa-3.1} model for multilingual multi-document English using a GPT-4o translated version of the Multi-News dataset~\cite{fabbri2019multi}. We fine-tune this model using Low-Rank Adaptation (LoRA)~\cite{hu2021lora} and utilize a 38,830 example training set of articles translated by GPT-4o. We utilize 7,726 and 7,521 test examples. We check the validation loss every 1000 steps and choose the model with the lowest validation loss. We use a LoRA rank of $r=16$ and $\alpha=32$, a dropout of 0.10, a learning rate of $2\times 10^{-4}$, and a batch size of 8 due to memory constraints. We utilize the following prompt to query \texttt{LLaMa-3.1} for a summary: \textit{You work for a news researcher and your job is to create an English summary of multilingual texts. Write a concise English-language summary of the following texts, where individual texts are separated by |||||: }

\begin{table}[h!]
\centering
\tiny
\selectfont
\setlength\tabcolsep{6pt}
\caption{ROUGE Scores}
\label{tab:rouge_scores}

\begin{tabular}{l|c|c|c|c}
 \toprule
\textbf{}   & \textbf{ROUGE-1} & \textbf{ROUGE-2} & \textbf{ROUGE-L} & \textbf{ROUGE-Lsum} \\ \midrule
{LLaMa 3.1}    & 0.3324           & 0.0953           & 0.1616           & 0.1881              \\ \hline
tuned-LLaMa 3.1 &  0.4138 &  0.1367 & 0.2180 & 0.2180 \\ \bottomrule
\end{tabular}

\end{table}

\begin{figure}
\noindent\fcolorbox{black}{lightgray}{
\begin{minipage}{.465\textwidth}
\scriptsize
```Washington (CNN) -- Yhdysvaltain Syyrian suurlähettiläs vieraili torstaina Haman taistelujen keskellä olevassa kaupungissa osana Yhdysvaltain tukea Syyrian demokratiataistelijoille. Suurlähettiläs Robert Ford vieraili Hamassa "tehdäkseen täysin selväksi fyysisellä läsnäolollaan, että tuemme niitä syyrialaisia, jotka ilmaisevat oikeuttaan puhua muutoksen puolesta, sanoi ulkoministeriön tiedottaja Victoria Nuland. Hamassa on ollut väkivaltaa ja yleislakko tällä viikolla sarjan rauhanomaisten mielenosoitusten jälkeen, mukaan lukien suuri hallituksen vastainen mielenosoitus viime perjantaina. Alueella seurasi raju tukahduttaminen, ja aktivistit sekä ihmisoikeusjärjestö Human Rights Watch raportoivat monista pidätyksistä ja kuolemista. Presidentti Bashar al-Assad erotti Haman maakunnan kuvernöörin lauantaina ja turvallisuusjoukot poistivat tankit kaupungin laitamille, mikä viittaa siihen, että jännitteet voisivat helpottua. Suurlähettiläs Ford tapasi yli tusinan verran Haman asukkaita ja vieraili sairaalassa, joka on hoitanut turvallisuusjoukkojen tukahduttamistoimien haavoittamia, Nuland sanoi ja lisäsi, että häntä tervehdittiin "erittäin lämpimästi". Valtiollinen uutistoimisto SANA raportoi ulkoministeriölähteen syyttäneen Fordia Hamaan menemisestä ilman hallituksen ennakkoon saamaa lupaa. Raportin mukaan ulkoministeriön virkailija sanoi, että Fordin vierailu oli "selvä todiste Yhdysvaltojen osallisuudesta käynnissä oleviin tapahtumiin Syyriassa ja sen pyrkimyksistä pahentaa tilanteita, jotka horjuttavat Syyriaa.  Nuland kuitenkin sanoi, että Yhdysvaltain viranomaiset ilmoittivat Syyrian hallitukselle, että suurlähetystön tiimi oli matkalla Hamaan. "Suurlähetystö ilmoitti Syy ||||| Reports of biggest crowd in Syria so far in city at heart of opposition, as activists say 13 dead across country.  More than 500,000 Syrians flooded through the city of Hama on Friday, according to activists, in what they claim was the single biggest protest yet against the embattled government of President Bashar al-Assad. The opposition reported 13 protesters killed, including five deaths in the central city of Homs, two in the capital\'s commercial neighbourhood Midan and six in the Dumair area, east of Damascus. Syrian state-run TV said the deaths in Damascus and Homs were caused by snipers from "armed gangs". An activist told Al Jazeera that Hama, where marchers were seen carrying olive branches, had become a "tangible example of resistance to injustice" in Syria. Hundreds of thousands also protested last Friday in Hama, prompting mass arrests and reports of several deaths when Syrian security forces stormed the city, Syria\'s third largest, and the surrounding area. "Hama, with all the support it is receiving from all over the country, is becoming a role model for peaceful demonstrations and we are protesting here for all of Syria," the local activist said. Western solidarity Friday\'s protests followed a visit to Hama by Robert Ford, the US ambassador to Syria, who toured the city on Thursday to show solidarity with residents, the US State Department said. Ford reached the city after passing checkpoints run by the military and Hama residents. A US official said Ford left Hama on Friday afternoon to avoid becoming a distraction during the weekly demonstrations. Diplomats said on Friday that French ambassador Eric Chevallier was also in Hama to show support. Damascus accused Washington of "interfering" in its affairs. "The presence of the US ambassador in Hama without previous permission is obvious proof of a clear evidence of the United States\' involvement in current events in Syria and its attempt to incite an escalation in the situation, which disturbs Syria\'s security and stability," the Syrian foreign ministry said in a statement. In response, the US state department said: "The fundamental intention was to make absolutely clear with his physical presence that we stand with those Syrians who are expressing their right to speak for change." In another development, Washington summoned the Syrian ambassador for talks this week after reports that embassy staff had filmed US protests against the crackdown in Syria. Ambassador Imad Mustapha was called in to meet with top State Department officials "to express a number of our concerns with the reported actions of certain Syrian embassy staff in the United States". Mustapha met with Assistant Secretary of State for Diplomatic Security Eric Boswell on Wednesday, according to a State Department statement. Fleeing Hama An estimated 1,000 Syrians have fled Hama in fear of another military crackdown on protests calling for Assad to quit and an end to the Baath Party\'s decades-long grip on power, a rights group said. The London-based Syrian Observatory for Human Rights said the residents had evacuated to Salamiyah, a town 30km from Hama, on Thursday. The rights group claimed that security forces have killed at least 23 civilians there and conducted mass arrests since Tuesday. 
\\\\

SUMMARY: The US has condemned the Syrian government for accusing its ambassador of interfering in the country\'s affairs after he visited the city of Hama yesterday, reports the BBC. The ambassador, Robert Ford, was in Hama to show support for protesters, says the State Department. "The fundamental intention was to make absolutely clear with his physical presence that we stand with those Syrians who are expressing their right to speak for change," said a State Department official. Meanwhile, reports Al Jazeera, hundreds of thousands of people marched in Hama today, the biggest protest yet against President Bashar al-Assad\'s government. Some 500,000 people marched through the city'
\end{minipage}
}
\caption{Example summary of multilingual documents. \label{fig:example-summary}}

\end{figure}

\clearpage
\onecolumn
\newpage
\section{Pearson Correlation on SemEval 2022 Task 8 Dataset
~\label{sec:app-pearson}}
\begin{table*}[!h]
\centering
\small
\caption{Comparison of different models' respective performances on the bilingual STS task. We bold the best score in each row. We find across nearly all language-combinations tests, as well as overall, that our multilingual-e5-base model fine-tuned using our modified AngIE~\cite{li2023angle} objective achieves the best results. }
\setlength\tabcolsep{6pt}
\resizebox{\textwidth}{!}{
\begin{tabular}{|c|l|c|c|c|c|c|c|c|c|}
\hline
\textbf{Dataset} & \textbf{Lang.} & \textbf{multi-e5} & \textbf{multi-e5-finetuned} & \textbf{multi-mpnet} & \textbf{multi-mpnet-finetuned} & \textbf{XLM-R}  & \textbf{XLM-R-finetuned} & \textbf{umt5-base}& \textbf{finetuned-umt5-base} \\ \hline
\multirow{6}{*}{\textbf{STS-22}} & en & 0.602 & 0.782& 0.512 & 0.767 &0.266 & 0.739& 0.192 & \textbf{0.789}\\ \cline{2-10} 
& de& 0.503 &\textbf{0.789}&0.324&0.724 &0.011 & 0.734 &0.017 & 0.783\\ \cline{2-10} 
& es & 0.679 & \textbf{0.835} & 0.528 &0.826 & 0.321 & 0.804 & 0.380 & 0.838\\ \cline{2-10} 
& zh & 0.701 & \textbf{0.777} & 0.647&0.767 &0.449 & 0.765& 0.466 & 0.767\\ \cline{2-10} 
& de-en& 0.586 & \textbf{0.693} & 0.548 &0.606 & 0.167 & 0.600 &0.067 & 0.673\\ \cline{2-10} 
& tr & 0.853 & \textbf{0.823} & 0.359 &0.776 & -0.119& 0.780 & 0.165 & 0.812\\ \cline{2-10} 

& pl & 0.602 & \textbf{0.776} & 0.404 &0.704 & 0.257 & 0.669& 0.221 &0.769 \\ \cline{2-10} 
& ar & 0.498 &\textbf{0.730} & 0.459&0.739 & 0.087 & 0.719& 0.210 & 0.722\\ \cline{2-10} 

& es-en & 0.698 &\textbf{0.859} & 0.597&0.842 & 0.328& 0.802 &0.302 & 0.847 \\ \cline{2-10}

& it & 0.732 & 0.882 & 0.489&0.862 & 0.370  & 0.821&  0.345 & \textbf{0.884}\\ \cline{2-10} 
& es-it & 0.615 & 0.866 & 0.405&0.827 & 0.331 & 0.802& 0.345 & \textbf{0.876}\\ \cline{2-10} 
& ru & 0.612 & \textbf{0.784} & 0.478& 0.762& 0.250 & 0.8745& 0.167 & 0.772\\ \cline{2-10} 
& fr & 0.722 & 0.898 & 0.479& 0.860& 0.228 & 0.863& 0.185 & \textbf{0.927}\\ \cline{2-10}

& de-fr & 0.466 & 0.803 & 0.417&0.623 & 0.041 & 0.681&0.127 & \textbf{0.873}\\ \cline{2-10} 
& pl-en & 0.739 & 0.818 & 0.826& \textbf{0.833} & 0.494 & 0.790&0.506 & 0.818  \\ \cline{2-10} 
& de-pl & 0.164 & 0.676 & 0.172&0.656 & -0.107 & 0.580& -0.021 & \textbf{0.871} \\ \cline{2-10} 
& fr-pl & 0.596& 0.687 & 0.521& {0.852} & 0.672  & 0.714& 0.289 &  \textbf{0.868}\\ \cline{2-10} 
& zh-en & 0.666 & \textbf{0.830} & 0.659&0.796& 0.336 & 0.754&0.285 & 0.794 \\ \cline{2-10} 
& all & 0.604  & \textbf{0.817} & 0.513 & 0.804& 0.226 & 0.749& 0.262 & 0.815 \\\bottomrule
\end{tabular}}
\end{table*}
\clearpage
\twocolumn
%\end{document}